\documentclass{article}

\usepackage{arxiv}

\usepackage{graphicx}
\usepackage[utf8]{inputenc} %
\usepackage[T1]{fontenc}    %
\usepackage{hyperref}       %
\usepackage{url}            %
\usepackage{booktabs}       %
\usepackage{amsfonts}       %
\usepackage{nicefrac}       %
\usepackage{microtype}      %
\usepackage{xcolor}
\usepackage{xspace}
\usepackage{wrapfig}
\usepackage{multirow}

\usepackage{amsmath}
\usepackage{amssymb}
\DeclareMathOperator*{\argmax}{arg\,max}

\def\etal{\emph{et al}.}
\def\ie{\emph{i}.\emph{e}.\ }
\def\eg{\emph{e}.\emph{g}.,\xspace}
\def\interpretGAN{InterpretGAN\xspace}

\newcommand{\tabincell}[2]{\begin{tabular}{@{}#1@{}}#2\end{tabular}}

\newcommand*{\red}{\textcolor{red}}

\title{
  Understanding and Diagnosing \\Vulnerability under Adversarial Attacks
}

\author{%
  Haizhong Zheng\\
  University of Michigan, Ann Arbor\\
  \texttt{hzzheng@umich.edu} \\
  \And
  Ziqi Zhang\\
  University of Michigan, Ann Arbor\\
  \texttt{ziqizh@umich.edu} \\
  \And
  Honglak Lee\\
  University of Michigan, Ann Arbor\\
  \texttt{honglak@eecs.umich.edu} \\
  \And
  Atul Prakash\\
  University of Michigan, Ann Arbor\\
  \texttt{aprakash@umich.edu} \\
}

\begin{document}

\maketitle

\begin{abstract}
Deep Neural Networks (DNNs) are known to be vulnerable to adversarial attacks.
Currently, there is no clear insight into how slight perturbations cause such a large difference in classification results and how we can design a more robust model architecture.
In this work, we propose a novel interpretability method, InterpretGAN, to generate explanations for features used for classification in latent variables.
Interpreting the classification process of adversarial examples exposes how adversarial perturbations influence features layer by layer as well as which features are modified by perturbations.
Moreover, we design the first diagnostic method to quantify the vulnerability contributed by each layer, which can be used to identify vulnerable parts of model architectures.
The diagnostic results show that the layers introducing more information loss tend to be more vulnerable than other layers.
Based on the findings, our evaluation results on MNIST and CIFAR10 datasets suggest that average pooling layers, with lower information loss, are more robust than max pooling layers for the network architectures studied in this paper.

\end{abstract}

\section{Introduction}
Deep Neural Networks (DNNs) have been found to be vulnerable to adversarial attacks~\cite{szegedy2013intriguing}.
Even slight imperceptible perturbations can lead to model misclassification.
To mitigate adversarial attacks, various types of defense methods~\cite{cohen2019certified,raghunathan2018certified,hein2017formal,wong2017provable,madry2017towards, zhang2019theoretically, tramer2017ensemble, zheng2019efficient} are proposed.
Besides building defense methods,
recent research also tries to interpret the reason behind the adversarial attack.
Ilyas \etal~\cite{ilyas2019adversarial} showed that DNNs tend to use non-robust features for classification, rather than robust features.
Another work~\cite{xu2019interpreting} applied network dissection~\cite{bau2017network} to interpret adversarial examples.
However, there is no clear understanding of the reason behind the vulnerability to adversarial attacks.
Since latent variables in DNNs are usually uninterpretable,
it is difficult to explain how adversarial perturbations influence features in latent variables and result in final misclassification.

If features influenced by perturbations can be translated to natural image space, we can understand which features are influenced in latent variables.
To achieve this goal, 
we propose a novel interpretability method, \interpretGAN, that can generate explanations for features relevant to classification in latent variables.
In \interpretGAN, we apply a Generative Adversarial Network (GAN) to train the explanation generator so that generated explanations follow a training data distribution and are interpretable. 
In order for explanations to contain the necessary features used for classification, we introduce an interpretability term for GAN training that ensures mutual information between explanations and classification results.
Whenever adversarial perturbations distort features used for classification, \interpretGAN can reflect those changes in the explanations.

We interpret the classification process of models trained on MNIST~\cite{lecun1998gradient} and CelebFaces~\cite{liu2015faceattributes} (with gender as the label) datasets
and use \interpretGAN to generate
explanations for each layer of the two models.
Based on generated explanations, we discover several interesting findings.
For classification of natural images, we find that as latent variables are processed layer by layer, features irrelevant to classification (e.g., hat for gender classification) are filtered out.
Furthermore, when adversarial examples are fed into the models, we notice that, rather than changing features directly in the input, \emph{adversarial perturbations influence latent variables gradually layer by layer},
and meaningless perturbations in the input cause semantic changes in latent variables.
For the CelebFaces model, perturbations change different kinds of features in different layers.
These explanations and findings can help researchers further understand the reason behind adversarial attacks.

A crucial problem of interpretability research is determining how to use interpretability results to improve model performance.
Bhatt \etal~\cite{bhatt2020explainable} conducted a survey showing that the largest demand for interpretability is to diagnose the model and to improve model performance.
In Section~\ref{sec:experiment-diagnose}, based on \interpretGAN, we propose the first diagnostic method that can quantify the vulnerability of each layer to help diagnose the problem of models layer by layer. This diagnostic method can be used for identifying vulnerable layers of model architectures and can provide insights to researchers towards designing more robust model architectures or adding additional regularizers to improve robustness of these layers.
We diagnose an MNIST model with the proposed method and find layers causing more information loss tend to be more vulnerable than other layers.
Based on the diagnostic results, we propose using average pooling layers that introduce less information loss than max pooling layers to improve the robustness.
Our evaluation results show that average pooling layers are more robust than max pooling layers on MNIST and CIFAR10~\cite{krizhevsky2009learning} datasets.

We summarize our contributions in this work as follows:
\begin{itemize}

    \item We propose a novel interpretability method, \interpretGAN, that generates explanations based on features relevant to classification in latent variables layer by layer. The generated explanations show how and what features in latent variables are influenced by adversarial perturbations layer by layer.

    \item To the best of our knowledge, our work is the first to propose a diagnostic method that quantifies the vulnerability contributed by each layer. This diagnostic method can help identify problems in model architectures with respect to robustness and can provide insights to researchers for building more robust models.

    \item Based on the insights offered by diagnostic results, the layers that introduce more information loss can be more vulnerable than other layers.
    We therefore conduct a study on the robustness of pooling layers. The evaluation results show that average pooling layers have better robustness compared to max pooling layers on both MNIST and CIFAR10 datasets.

\end{itemize}

\vspace{-0.3cm}

\section{Preliminaries}

\subsection{Adversarial Examples and Adversarial Training}

Szegedy \etal~\cite{szegedy2013intriguing} showed that small adversarial perturbations $\delta$ on an input $x$, which lead to a misclassification by deep neural networks, can be found by solving the following:
$$\delta = \argmax_{\delta \in \mathcal{S}} L(f(x + \delta), y),$$
where $\mathcal{S}$ is the allowed perturbation space, $L$ is  the loss function, $f$ is the classification model, and $y$ is the ground truth label.
Defending against adversarial attacks is an active research area~\cite{cohen2019certified,raghunathan2018certified,hein2017formal,wong2017provable, zheng2019efficient, shafahi2019adversarial}, and adversarial training~\cite{madry2017towards}, which formulates training as a game between a classification model and adversarial attacks, is one of the most effective defense methods.
To get clues on how to design a more robust model architecture for adversarial training,
we propose a model architecture diagnostic method to quantify the vulnerability contributed by each layer in Section~\ref{sec:experiment-diagnose}.

\subsection{Generative Adversarial Network}
GAN was introduced in \cite{goodfellow2014generative} as a framework for training generative models.
GAN has a generator and a discriminator network.
The generator $G$ takes a latent variable $z \sim p_z$ following a prior distribution, and its output $G(z)$ matches the real data distribution $p_{data}$.
$G$ is trained against an adversarial discriminator $D$ that tries to distinguish $G(z)$ from the data distribution $p_{data}$. The min-max game of GAN training can be formalized by:%

\begin{equation}\label{eq:gan}
\min_G \max_D V(D,G) = \mathbb{E}_{x\sim p_{data}}[\log D(x)] + \mathbb{E}_{z \sim p_z}[\log(1-D(G(z)))].
\end{equation}

Chen \etal~\cite{chen2016infogan}  found that it is possible to generate images with certain properties by adding corresponding regularizers in the loss function.
In this work,
we use GAN to train a generator that generates explanations that contain the necessary information of features used for classification in latent variables.

\section{Methodology: InterpretGAN}
An intriguing property of adversarial examples is that slight perturbations in input space can lead to large changes in output labels.
Since latent variables of DNNs are usually uninterpretable, it is difficult to figure out how adversarial perturbations influence features in latent variables.
If we can invert modified features back to images, we can understand how adversarial perturbations influence these features.
To achieve this goal, we propose a novel interpretability method, \interpretGAN,
that generates explanations for latent variables based on features relevant to classification.

\begin{figure}[htbp]
  \centering
  \begin{minipage}[t]{0.47\textwidth}
    \centering
    \includegraphics[width=1\textwidth]{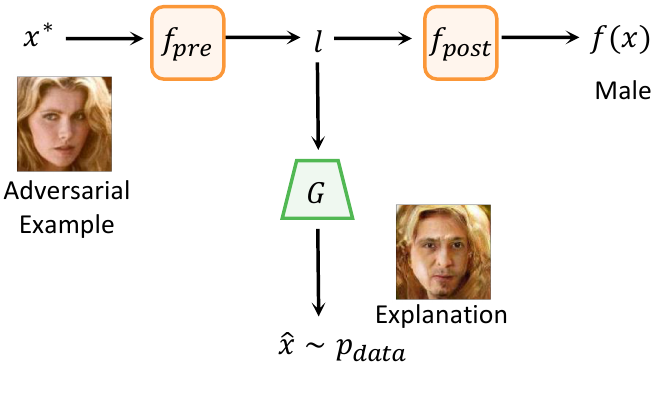}
    \caption{The network $G$ generates explanations $\hat{x}$ based on features relevant to classification in latent variable $l$, and $\hat{x}$ follows data distribution $p_{data}$.}
    \label{fig:interGAN-basic}
  \end{minipage}
  \hspace{0.5cm}
  \begin{minipage}[t]{0.47\textwidth}
    \centering
    \includegraphics[width=1\textwidth]{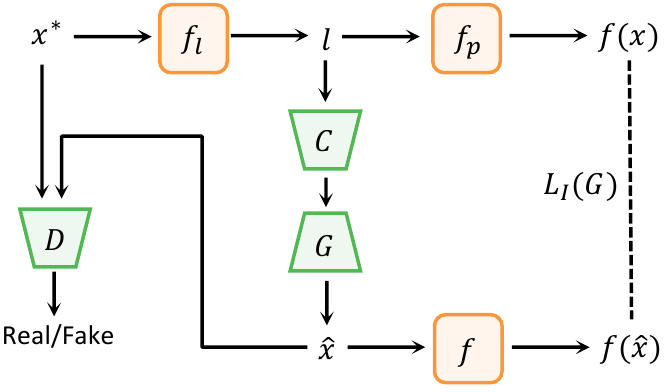}
    \caption{Training framework of \interpretGAN. Compression network $C$ can reduce the dimension of latent variables. $L_I(G)$ is the interpretability term.}
    \label{fig:training-framework}
  \end{minipage}
\end{figure}

The proposed \interpretGAN method uses GAN to train an explanation generator per layer that
generates explanations for relevant features in that layer to images following distribution $p_{data}$.
For a classification model $f$ and a latent variable $l$ of a layer, we denote the composition of layers before $l$ as $f_{pre}$ and composition of layers after $l$ as $f_{post}$, \ie, $f = f_{post} \circ f_{pre}$, and $l = f_{pre}(x)$.
As shown in Figure~\ref{fig:interGAN-basic}, we treat the visual explanation of latent variable $l$ as a generative process with interpretability regularization.
Instead of reconstructing original images from latent variables~\cite{mahendran2015understanding, dosovitskiy2016inverting},
\interpretGAN generates explanations that only contain necessary features relevant to the classification. Generated explanations are thus expected to differ from the original images since unnecessary features may be discarded.

To sum up, explanation $\hat{x}$ needs to satisfy two properties:
\textbf{1)} $\hat{x}$ should follow the data distribution $p_{data}$ to help guarantee that it is interpretable,
and \textbf{2)} $\hat{x}$ captures the necessary information in the latent variables used for classification.

Since we use GAN to train network $G$, the generated explanations naturally satisfy the first property~\cite{goodfellow2014generative}.
For the second property, we add an \emph{interpretability term} in the min-max game of GAN (Eq.~\ref{eq:interpretGAN-loss}).
The objective of the interpretability term is for the generated explanation $\hat{x}$ to contain a high mutual information with the classification result $f(x)$, which is $I(\hat{x}, f(x))$.
The higher the value of $I(\hat{x}, f(x))$, the more information used for classification is contained in $\hat{x}$.

Therefore, to train \interpretGAN, we formalize the min-max game for our GAN training as the following, including a mutual information term as compared to equation~\ref{eq:gan}:
\begin{equation} \label{eq:interpretGAN-loss}
  \begin{split}
    \min_G \max_D V_I(D,G)
    &= V(D,G) - \lambda I(\hat{x}, f(x))\\
    &= V(D,G) - \lambda I(G(l), f(x)),\\
    \text{where}\ V(D,G) &= \mathbb{E}_{x\sim p_{data}}[\log D(x)] + \mathbb{E}_{l \sim p_l}[\log(1-D(G(l)))].
  \end{split}
\end{equation}

The calculation of the interpretability term $I(G(l), f(x))$ is intractable during training, but we can maximize the variational lower bound for $I(G(l), f(x))$ as a surrogate\footnote{The derivation process can be found in the supplementary material.}:

\begin{equation}
  \begin{split}
    I(G(l), f(x))
    &=I(\hat{x}, y)
    =H(y) - H(y|\hat{x})\\
    &\geq \int_{\hat{x}} p(\hat{x}) \int_y p(y|\hat{x}) \log {p(\hat{y}|\hat{x})} dy d\hat{x} + H(y)\\
    &= \mathbf{E}_{\hat{x}\sim G(l),y \sim f(x)}[\log {p(\hat{y}|\hat{x})}] + H(y) \triangleq L_I(G),
  \end{split}
\end{equation}
where $y$ is $f(x)$, and $\hat{y}$ is $f(\hat{x})$. So the final optimization problem becomes:
\begin{equation}
\min_G \max_D V_I(D,G) = V(D,G) - \lambda L_I(G).
\end{equation}

The training framework of our model is illustrated in Figure~\ref{fig:training-framework}. In practice, since some latent variables have a large dimension, we adopt an extra compression network $C$ between the latent variables and the generative model. $C$ compresses the latent variables and controls the dimension of input for the generative model, which causes GAN training converge more efficiently.

In the following two sections, we divide our evaluation into two parts. In Section~\ref{sec:experiment-interpret}, \interpretGAN is used to explain how models classify natural images and adversarial images, and how adversarial perturbations lead to misclassification.
In Section~\ref{sec:experiment-diagnose}, a diagnostic method is designed based on \interpretGAN to quantify the vulnerability of each layer in a model.

\section{Visual Explanation of Latent Variables for Classification}\label{sec:experiment-interpret}

In this section, we apply our method to interpret classification models trained with two datasets: MNIST~\cite{lecun1998gradient} and CelebFaces ~\cite{liu2015faceattributes} with gender as the label.
For MNIST, we resize all images to $32 \times 32$ for the convenience of GAN training and use a widely used model architecture in adversarial training research~\cite{madry2017towards, zhang2019theoretically, shafahi2019adversarial, zhang2019you, zheng2019efficient}.  We use VGG$16$~\cite{simonyan2014very} for the CelebFaces classification,
and the generative models are trained with progressive GAN~\cite{karras2017progressive}.
For MNIST, we interpret outputs for all convolutional and fully connected layers.
For CelebFaces, we interpret outputs of all convolutional blocks and fully connected layers.

The results of interpreting the classification process of natural images and adversarial examples for the naturally trained models are presented in Section~\ref{ssec:interpret-adv}. Due to space limits, additional explanation results and detailed experiment configuration can be found in the supplementary material.

\begin{figure}[!t]
  \centering
  \includegraphics[width=0.9\textwidth]{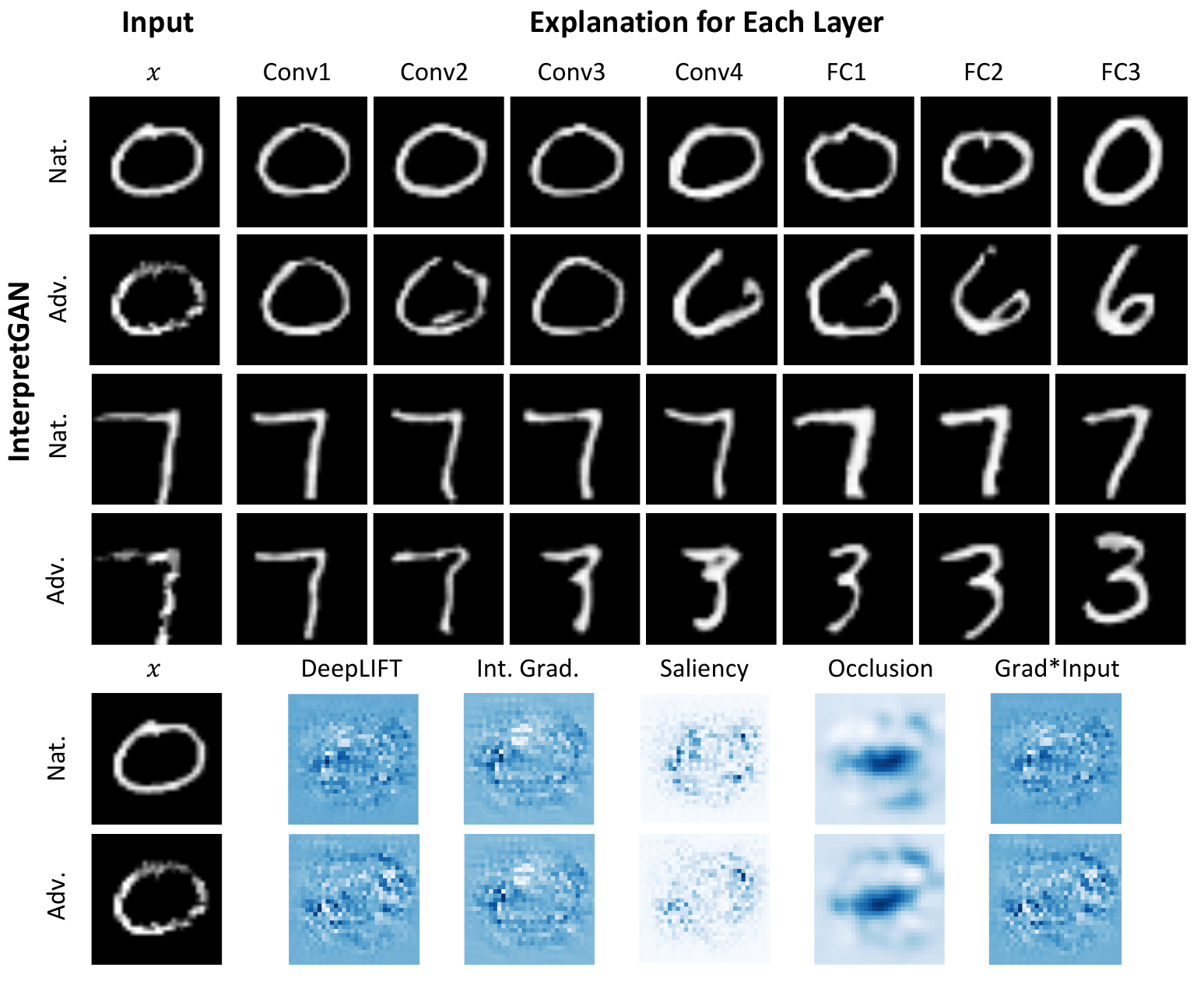}
  \caption{
 Explanations of classifications on natural images and corresponding adversarial images on the MNIST dataset. Explanations of adversarial examples are changed gradually through layers. 
Other interpretability methods can show the importance of pixels for classification of the natural image, but the explanation for the adversarial examples can't explain the reason behind misclassification.
  }
  \label{fig:adv-interpret-mnist}
\end{figure}

\begin{figure}[!htbp]
  \centering
  \includegraphics[width=0.97\textwidth]{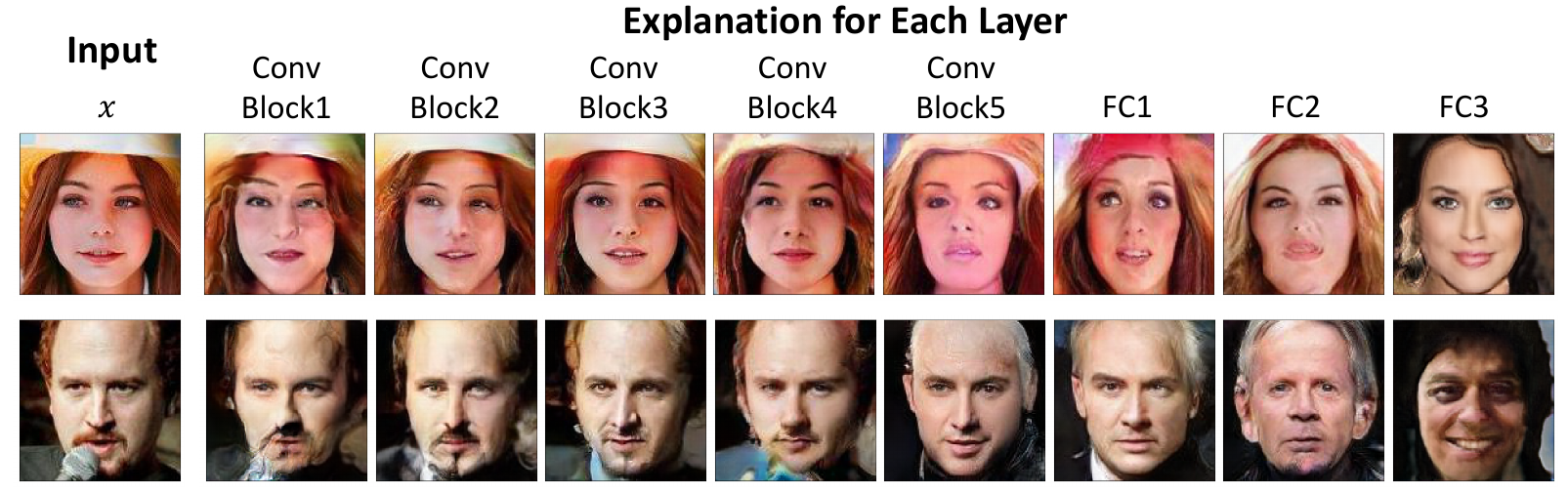}
  \caption{ Generated explanations with features irrelevant to gender classification. The woman's hat in the first row is transformed into her hair during classification. The microphone in the second row is filtered in the first two convolutional blocks and the blocked face is completed by the model.
    }
  \label{fig:nat-interpret-celeb}
\end{figure}

\subsection{Interpreting Classification Process}\label{ssec:interpret-adv}
\begin{figure}[!t]
  \centering
  \includegraphics[width=0.92\textwidth]{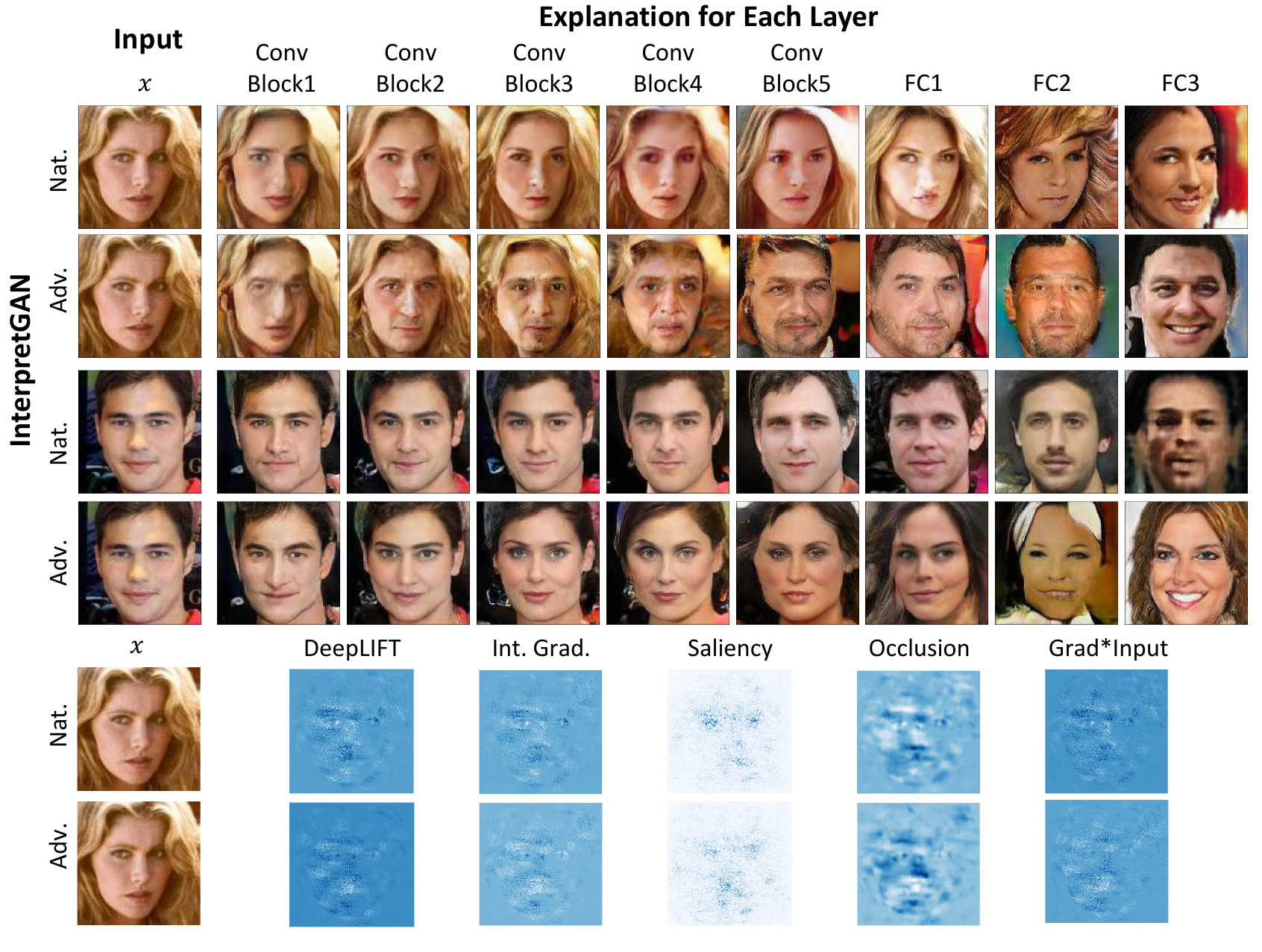}
  \caption{ Explanations generated by our method and other baseline methods. InterpretGAN can illustrate which features are modified by adversarial perturbations. In shallower layers, local facial features are modified and, in deeper layers, backgrounds and hair are modified.
  Other interpretability methods have similar results to explanations on the MNIST dataset.
    }
  \label{fig:adv-interpret-celeb}
\end{figure}

We interpret two models naturally trained with MNIST and CelebFaces datasets
and compare the explanation results of our method to other interpretability methods (DeepLIFT~\cite{shrikumar2017learning}, saliency maps~\cite{simonyan2013deep}, input$\times$gradient~\cite{kindermans2016investigating}, integrated gradient~\cite{sundararajan2017axiomatic}, and occlusion~\cite{zeiler2014visualizing}),
which explain the relation between classification results and features.
To investigate how adversarial perturbations influence features in the MNIST dataset, we use PGD-$40$ ($l_\infty, \epsilon = 0.3$)~\cite{kurakin2016adversarial} and PGD-$20$ ($l_\infty, \epsilon = 2/255$) untargeted attacks to generate adversarial examples for MNIST and CelebFaces models. 
Figures~\ref{fig:adv-interpret-mnist} and~\ref{fig:adv-interpret-celeb} show the generated explanations for MNIST model and CelebFaces model, respectively.

\textbf{Classification of Natural Images.}
The first and third rows in Figures~\ref{fig:adv-interpret-mnist} and ~\ref{fig:adv-interpret-celeb} present the explanations of natural image classification.
Since the generated explanations are regularized with the interpretability term in Eq.~\ref{eq:interpretGAN-loss}, 
we find that the explanations for natural images are consistent with the classification results, but not identical to the inputs\footnote{In FC layers, dimensions of latent variables are largely compressed, and features lose detailed information, which makes it harder for $G$ to invert details and causes the large changes in FC layers.}.
It is also observed that the model tries to only keep necessary features used for classification and tends to drop irrelevant features in latent variables.
As shown in Figure~\ref{fig:nat-interpret-celeb}, for a woman with a hat and a man with a microphone,
the irrelevant features (hat and microphone) are gradually removed or transformed layer by layer.

\textbf{Classification of Adversarial Examples.} When we feed adversarial examples into the models, the explanations are quite different. The second and fourth rows in Figures~\ref{fig:adv-interpret-mnist} and~\ref{fig:adv-interpret-celeb} show the explanations for corresponding adversarial examples.
One interesting finding we notice for both models is that, rather than changing features directly in adversarial examples, \emph{adversarial perturbations influence latent variables gradually layer by layer}: the deeper the layer is,  the closer the explanation is to the misclassified label. 
However, we cannot find such difference in explanations generated by other baseline methods.
We also find that adversarial perturbations influence features in latent variables in different ways for the MNIST and CelebFaces models.

For the MNIST model, although adversarial perturbations only change pixel values in the input images,
explanations show that perturbations actually also modify features related to pixel position in latent variables.
Figure~\ref{fig:adv-interpret-mnist} shows that the process of misclassification of adversarial "$0$" and "$7$".
In addition to pixel values, pixel positions are also gradually changed
so that perturbations can reuse these pixels to form new features for misclassified labels.
The explanation "$6$" for adversarial "$0$" reuses most parts of original "$0$", and the explanation "$3$" for adversarial "$7$" is the curved version of the original "$7$".

When we perform the untargeted attack on MNIST,
we find that adversarial examples with the same ground truth label are more likely to be misclassified as a certain label.
For example, with the untargeted attack, $39.3\%$ adversarial "$0$"s are misclassified as "$6$", and $39.5\%$ adversarial "$7$"s are misclassified as "$3$".\footnote{More results can be found in the supplementary material.}
We think the reason is that it is easier for perturbations to spatially move original pixels to form features for the misclassified labels compared to only changing pixels values.

The CelebFaces dataset shows that adversarial perturbations influence different types of features in different layers: local features (e.g., nose, eyes, mouth, and beard) are more likely to be changed in shallower layers, and global features (e.g., hair and background) tend to be changed in deeper layers.
As shown in Figure~\ref{fig:adv-interpret-celeb}, when an adversarial "female" is fed into the model, the perturbations change eyes and nose first and then add a beard to the face. 
At a latter stage, the long blond hair is changed to short black hair. For the adversarial examples in the fourth line, eyes, lip, and make-up are changed in the shallower layers. 
Hair becomes longer in the explanations of Conv Block4 and Conv Block5.

\section{Diagnosing Model Architecture}\label{sec:experiment-diagnose}
A crucial need for interpretability methods is how to take advantage of interpretability results to improve model performance~\cite{bhatt2020explainable}.
What we have presented so far provides visual explanations for adversarial attacks, but not a diagnostic procedure to improve the model. In this section, we design a novel diagnostic method that can quantify the vulnerability contributed by each layer to adversarial attacks. We accomplish that by feeding the generated explanations back into the classification model and developing a metric to measure vulnerability contributed by each layer, as outlined next.

\subsection{Quantifying Vulnerability for Layers}
For a layer whose output is latent variable $l$ (\ie, $l=f_{pre}(x)$, $f = f_{post} \circ f_{pre}$)
and its generated explanation $\hat{x} = G(C(f_{pre}(x)))$,
since features of $\hat{x}$ are only influenced by perturbations related to $f_{pre}$
and $\hat{x}$ follows the natural data distribution $p_{data}$,
perturbations related to $f_{post}$ are filtered out in $\hat{x}$.
If we feed $\hat{x}$ back to the classification model $f$,
we can measure how much adversarial perturbations related to $f_{pre}$ change the classification result.

To evaluate the vulnerability of each layer against an adversarial dataset $\mathcal{D^*}$,  we calculate the accuracy $Acc_i$ of the generated explanation $\hat{x}$ for the $i$th layer and introduce a metric defined as follows: the vulnerability ${Vul}_i$ of the $i$th layer is the accuracy drop as compared to the previous layer. Mathematically, it is formulated as:
\vspace{-0.1cm}
\begin{equation}
  \begin{split}
    Vul_i &=
    \begin{cases}
      1 - Acc_i & i=1 \\
      Acc_{i-1} - Acc_i & i>1
   \end{cases}\\
   Acc_i &= \mathbb{E}_{(x^*,y) \in D^*}[\mathbf{1}_{\{true\}}(f(\hat{x_i}) = y) ]\\
   \hat{x}_i &= G(C(f_{pre_i}(x^*))),
  \end{split}
\end{equation}

where $\mathbf{1}$ is the indicator function, and $f_{{pre}_i}$ is the composition of layers until the $i$th layer.
\subsection{Example: Diagnosis of MNIST Models}
\begin{table}[!htbp]
  \centering
  \begin{tabular}{ccccccccc}
  \toprule
  \multicolumn{2}{c}{Layers}  & Conv1   & Conv2  & Conv3   & Conv4  & FC1     & FC2     & FC3    \\ \midrule
  \multirow{2}{*}[-0em]{\begin{tabular}[c]{@{}c@{}}Naturally \\Trained\end{tabular}} & $Acc_i$ & $52.6\%$  & $26.3\%$ & $21.6\%$ & $9.9\%$  & $4.3\%$  & $3.0\%$  & $3.5\%$ \\ %
& $Vul_i$ &\red{$47.4\%$}&\red{$26.3\%$}&$4.7\%$&\red{$11.7\%$}&\red{$5.6\%$}&$1.3\%$&$-0.5\%$\\ \midrule
  \multirow{2}{*}[-0em]{\begin{tabular}[c]{@{}c@{}}Adversarially\\Trained \end{tabular}} & $Acc_i$ & $95.1\%$ & $94.0\%$ & $93.9\%$  & $92.6\%$ & $90.1\%$ & $90.0\%$ & $89.5\%$ \\ %
& $Vul_i$ &\red{$4.9\%$}&\red{$1.1\%$}&$0.1\%$&\red{$1.3\%$}&\red{$2.5\%$}&$0.1\%$&$0.5\%$\\
  \bottomrule
  \end{tabular}
  \vspace{0.1cm}
  \caption{Accuracy on \interpretGAN-generated images and the vulnerability of each layer for naturally- and adversarially-trained MNIST models. Red numbers indicate large accuracy drop (vulnerability).}
  \label{tab:diagnose-mnist}
\end{table}

\begin{figure}[!htbp]
  \centering
  \includegraphics[width=0.97\textwidth]{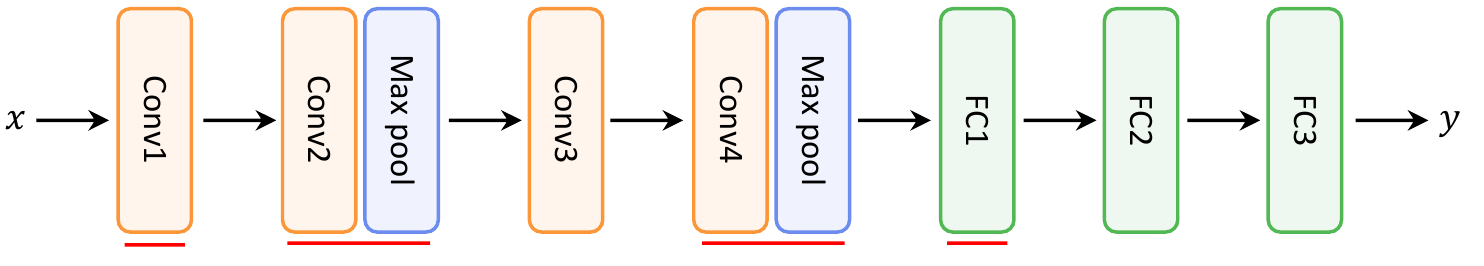}
  \caption{Architecture of the model used for MNIST classification. Red lines indicate layers with large accuracy drop (vulnerability) on generated images.}
  \label{fig:mnist-architecture}
\end{figure}
We use the proposed method to quantify the vulnerability of MNIST model architecture used in Section~\ref{sec:experiment-interpret}
and then quantify the vulnerability for a naturally trained model and an adversarially trained model.
We perform PGD-$40$ $l_\infty$($\epsilon=0.3$) attack on a test set to obtain $D^*$, and we use ATTA-$1$~\cite{zheng2019efficient} to adversarially train the models.

Table~\ref{tab:diagnose-mnist} shows the diagnostic results.
For both models, accuracy drops in almost all layers, which is consistent to our finding that perturbations influence latent variables gradually through layers\footnote{The naturally trained model has a negative vulnerability in layer FC3. This fluctuation is believed to be caused by that output distribution of model $G$ just approximates $p_{data}$ rather than being identical to $p_{data}$.}.
Specifically, we find that four layers (Conv1, Conv2, Conv4, and FC1) are more vulnerable than other layers.
To study the reason behind these vulnerable layers, we locate them in the model architecture shown in Figure~\ref{fig:mnist-architecture}.
For Conv1 (Conv layer after image) and FC1 (FC layer after Conv layer), we think the reason behind high vulnerability is that these two layers change the format of latent variables, which causes a higher information loss. Conv2 and Conv4 contain max pooling layers that down-sample latent features.
The pooling layers in the MNIST model use $2\times 2$ filter and $2\times 2$ stride, which drops $75\%$ of latent variables during classification.
The higher information loss can be the cause for the vulnerability of these layers.
We compare the difference of max pooling indices (positions of the maximum value) between natural images and adversarial examples
and find that, with PGD-$40$ ($l_\infty, \epsilon = 0.3$) attack,
$40.9\%$ and $18.1\%$ max pooling indices change in the first and second pooling layers, respectively, which means that some important features in maximum variables used for natural classification are dropped.
Besides changing the value of features, adversarial perturbations also drop important features in max pooling layers, likely leading to higher vulnerability of these layers.

\begin{table}[!htbp]
  \centering
  \begin{tabular}{cccccc}
  \toprule
              & PGD-$40$   & PGD-$100$  & M-PGD-$40$   & FGSM    \\ \midrule
  Max Pooling & $89.33\%$  & $81.98\%$ & $89.76\%$ & $95.62\%$  \\
  Avg Pooling & $91.56\%$  & $87.48\%$ & $92.09\%$ & $95.72\%$  \\
  \bottomrule
  \end{tabular}
  \caption{Comparison of robustness between max pooling model and average pooling model on the MNIST. Model with average pooling layers show better robustness.}
  \label{tab:robustness-mnist}
\end{table}

\begin{table}[!htbp]
  \centering
  \begin{tabular}{cccccccc}
  \toprule
  &             & PGD-$20$   & PGD-$100$  & M-PGD-$20$  & FGSM    \\ \midrule
  \multirow{2}{*}[-0em]{\begin{tabular}[c]{@{}c@{}}VGG16\end{tabular}}
  & Max Pooling & $46.97\%$  & $45.69\%$ & $47.06\%$ & $50.46\%$  \\
  & Avg Pooling & $48.25\%$  & $47.09\%$ & $48.91\%$ & $51.64\%$  \\
  \midrule

  \multirow{2}{*}[-0em]{\begin{tabular}[c]{@{}c@{}}WideResNet\end{tabular}}
  & Max pooling & $53.08\%$  & $51.65\%$ & $53.94\%$ & $57.63\%$  \\
  & Avg pooling & $54.68\%$  & $52.59\%$ & $55.47\%$ & $59.92\%$  \\

  \bottomrule
  \end{tabular}
  \caption{
With average pooling layers, CIFAR10 models show better robustness under different adversarial attacks.
}
  \label{tab:robustness-cifar}
\end{table}

To mitigate the information loss in the max pooling layer, we propose using average pooling instead of max pooling. Compared to max pooling, average pooling uses all features to calculate outputs, which can retain more information during down-sampling.
We adversarially retrain the MNIST model with average pooling layers and present a comparison of robustness under different attacks in Table~\ref{tab:robustness-mnist}.
The results show that model robustness can be improved by simply replacing the max pooling layer with the average pooling layer.
We also evaluate this experiment on the CIFAR10 dataset with VGG16~\cite{simonyan2014very} and WideResNet~\cite{zagoruyko2016wide}. The evaluation results shown in Table~\ref{tab:robustness-cifar} also indicate that the average pooling layer tends to be more robust than the max pooling layer in these networks.

\section{Related Work}
\textbf{Interpretability methods.}
Lots of efforts have tried understanding how DNNs classify images.
Some works~\cite{ribeiro2016should} study the relation between classification results and input features and
most of the works~\cite{kindermans2016investigating, sundararajan2017axiomatic, zeiler2014visualizing} focus on estimating the importance (\eg Shapley Value) of pixels in images.
Others~\cite{mahendran2015understanding, dosovitskiy2016inverting, selvaraju2017grad} also try to invert latent features back to the image space, but unlike our work, which generates explanation only based on relevant features used for classification, these works focus on reconstructing complete input images from all features encoded in latent variables.
Similar to our work, there are also some recent research~\cite{singla2019explanation, samangouei2018explaingan, dosovitskiy2016generating} designing interpretability methods based on generative models. The generative models generate explanations to determine which features need to be changed in input to change the confidence of classification. Singla \etal~\cite{singla2019explanation} also uses this method to generate saliency maps for input images. Unlike our work, none of the previous works study how adversarial perturbations influence features in latent variables.

\textbf{Insights for building robust models.}
The reasons behind the existence of adversarial examples and how to build more robust models are still open questions.
Goodfellow \etal~\cite{goodfellow2014explaining, zhou2016learning} suggested the cause of adversarial examples is the linear nature of layers in neural networks with sufficient dimensionality. Ilyas \etal~\cite{ilyas2019adversarial} found that neural networks tend to use non-robust features for classification rather than robust features. Other works~\cite{zhang2019interpreting, etmann2019connection} studied why adversarially trained models are more robust than naturally trained models. Xu \etal~\cite{xu2019interpreting} applied network dissection~\cite{bau2017network} to interpret adversarial examples. Unlike previous research, in this work, we interpret adversarial examples from a new perspective, which produce additional insights for researchers to understand adversarial attacks.

\vspace{-0.3cm}
\section{Conclusion}
\vspace{-0.2cm}

In this work, we propose a novel interpretability method, as well as the first diagnostic method that can quantify the vulnerability contributed by each layer against adversarial attacks.
By factoring in the mutual information between generated images and classification results, \interpretGAN generates explanations containing relevant features used for classification in latent variables at each layer. The generated explanations show how and what features are influenced by adversarial perturbations.
We also propose a new vulnerability metric that can be computed using \interpretGAN to provide insights to identify vulnerable parts of models against adversarial attacks.
Based on the diagnostic results, we study the relationship between robustness and model architectures.
Specifically, we find that, to design a robust model, average pooling layers can be a more reliable than max pooling layers.

\textbf{Acknowledgement.} This material is based on work supported by DARPA under agreement number 885000,  the National Science Foundation (NSF) Grants 1646392, and Michigan Institute of Data Science (MIDAS).   We also thank researchers at Ford for their feedback on this work and support from Ford to the University of Michigan.

\small

{\small
\bibliographystyle{neurips_2020}
\bibliography{egbib}

\begin{thebibliography}{10}

\bibitem{szegedy2013intriguing}
Szegedy, C., W.~Zaremba, I.~Sutskever, et~al.
\newblock Intriguing properties of neural networks.
\newblock In \emph{International Conference on Learning Representations
  (ICLR)}. 2014.

\bibitem{cohen2019certified}
Cohen, J.~M., E.~Rosenfeld, J.~Z. Kolter.
\newblock Certified adversarial robustness via randomized smoothing.
\newblock In \emph{International Conference on Machine Learning (ICML)}. 2019.

\bibitem{raghunathan2018certified}
Raghunathan, A., J.~Steinhardt, P.~Liang.
\newblock Certified defenses against adversarial examples.
\newblock In \emph{International Conference on Learning Representations
  (ICLR)}. 2018.

\bibitem{hein2017formal}
Hein, M., M.~Andriushchenko.
\newblock Formal guarantees on the robustness of a classifier against
  adversarial manipulation.
\newblock In \emph{Advances in Neural Information Processing Systems
  (NeurIPS)}, pages 2266--2276. 2017.

\bibitem{wong2017provable}
Wong, E., J.~Z. Kolter.
\newblock Provable defenses against adversarial examples via the convex outer
  adversarial polytope.
\newblock In \emph{International Conference on Machine Learning (ICML)}. 2018.

\bibitem{madry2017towards}
Madry, A., A.~Makelov, L.~Schmidt, et~al.
\newblock Towards deep learning models resistant to adversarial attacks.
\newblock In \emph{International Conference on Learning Representations
  (ICLR)}. 2018.

\bibitem{zhang2019theoretically}
Zhang, H., Y.~Yu, J.~Jiao, et~al.
\newblock Theoretically principled trade-off between robustness and accuracy.
\newblock In \emph{International Conference on Machine Learning (ICML)}. 2019.

\bibitem{tramer2017ensemble}
Tram{\`e}r, F., A.~Kurakin, N.~Papernot, et~al.
\newblock Ensemble adversarial training: Attacks and defenses.
\newblock In \emph{International Conference on Learning Representations
  (ICLR)}. 2018.

\bibitem{zheng2019efficient}
Zheng, H., Z.~Zhang, J.~Gu, et~al.
\newblock Efficient adversarial training with transferable adversarial
  examples.
\newblock \emph{arXiv preprint arXiv:1912.11969}, 2019.

\bibitem{ilyas2019adversarial}
Ilyas, A., S.~Santurkar, D.~Tsipras, et~al.
\newblock Adversarial examples are not bugs, they are features.
\newblock \emph{arXiv preprint arXiv:1905.02175}, 2019.

\bibitem{xu2019interpreting}
Xu, K., S.~Liu, G.~Zhang, et~al.
\newblock Interpreting adversarial examples by activation promotion and
  suppression.
\newblock \emph{arXiv preprint arXiv:1904.02057}, 2019.

\bibitem{bau2017network}
Bau, D., B.~Zhou, A.~Khosla, et~al.
\newblock Network dissection: Quantifying interpretability of deep visual
  representations.
\newblock In \emph{Proceedings of the IEEE Conference on Computer Vision and
  Pattern Recognition (CVPR)}, pages 6541--6549. 2017.

\bibitem{lecun1998gradient}
LeCun, Y., L.~Bottou, Y.~Bengio, et~al.
\newblock Gradient-based learning applied to document recognition.
\newblock \emph{Proceedings of the IEEE}, 86(11):2278--2324, 1998.

\bibitem{liu2015faceattributes}
Liu, Z., P.~Luo, X.~Wang, et~al.
\newblock Deep learning face attributes in the wild.
\newblock In \emph{Proceedings of International Conference on Computer Vision
  (ICCV)}. 2015.

\bibitem{bhatt2020explainable}
Bhatt, U., A.~Xiang, S.~Sharma, et~al.
\newblock Explainable machine learning in deployment.
\newblock In \emph{Proceedings of the 2020 Conference on Fairness,
  Accountability, and Transparency}, pages 648--657. 2020.

\bibitem{krizhevsky2009learning}
Krizhevsky, A., G.~Hinton, et~al.
\newblock Learning multiple layers of features from tiny images.
\newblock Tech. rep., Citeseer, 2009.

\bibitem{shafahi2019adversarial}
Shafahi, A., M.~Najibi, A.~Ghiasi, et~al.
\newblock Adversarial training for free!
\newblock In \emph{Advances in Neural Information Processing Systems
  (NeurIPS)}. 2019.

\bibitem{goodfellow2014generative}
Goodfellow, I., J.~Pouget-Abadie, M.~Mirza, et~al.
\newblock Generative adversarial nets.
\newblock In \emph{Advances in Neural Information Processing Systems
  (NeurIPS)}, pages 2672--2680. 2014.

\bibitem{chen2016infogan}
Chen, X., Y.~Duan, R.~Houthooft, et~al.
\newblock Infogan: Interpretable representation learning by information
  maximizing generative adversarial nets.
\newblock In \emph{Advances in Neural Information Processing Systems
  (NeurIPS)}, pages 2172--2180. 2016.

\bibitem{mahendran2015understanding}
Mahendran, A., A.~Vedaldi.
\newblock Understanding deep image representations by inverting them.
\newblock In \emph{Proceedings of the IEEE Conference on Computer Vision and
  Pattern Recognition (CVPR)}, pages 5188--5196. 2015.

\bibitem{dosovitskiy2016inverting}
Dosovitskiy, A., T.~Brox.
\newblock Inverting visual representations with convolutional networks.
\newblock In \emph{PProceedings of the IEEE Conference on Computer Vision and
  Pattern Recognition (CVPR)}, pages 4829--4837. 2016.

\bibitem{zhang2019you}
Zhang, D., T.~Zhang, Y.~Lu, et~al.
\newblock You only propagate once: Accelerating adversarial training via
  maximal principle.
\newblock In \emph{Advances in Neural Information Processing Systems
  (NeurIPS)}. 2019.

\bibitem{simonyan2014very}
Simonyan, K., A.~Zisserman.
\newblock Very deep convolutional networks for large-scale image recognition.
\newblock In \emph{International Conference on Learning Representations
  (ICLR)}. 2014.

\bibitem{karras2017progressive}
Karras, T., T.~Aila, S.~Laine, et~al.
\newblock Progressive growing of gans for improved quality, stability, and
  variation.
\newblock In \emph{International Conference on Learning Representations
  (ICLR)}. 2018.

\bibitem{shrikumar2017learning}
Shrikumar, A., P.~Greenside, A.~Kundaje.
\newblock Learning important features through propagating activation
  differences.
\newblock In \emph{International Conference on Machine Learning (ICML)}, pages
  3145--3153. JMLR. org, 2017.

\bibitem{simonyan2013deep}
Simonyan, K., A.~Vedaldi, A.~Zisserman.
\newblock Deep inside convolutional networks: Visualising image classification
  models and saliency maps.
\newblock In \emph{International Conference on Learning Representations
  (ICLR)}. 2014.

\bibitem{kindermans2016investigating}
Kindermans, P.-J., K.~Sch{\"u}tt, K.-R. M{\"u}ller, et~al.
\newblock Investigating the influence of noise and distractors on the
  interpretation of neural networks.
\newblock In \emph{Advances in Neural Information Processing Systems
  (NeurIPS)}. 2016.

\bibitem{sundararajan2017axiomatic}
Sundararajan, M., A.~Taly, Q.~Yan.
\newblock Axiomatic attribution for deep networks.
\newblock In \emph{International Conference on Machine Learning (ICML)}, pages
  3319--3328. 2017.

\bibitem{zeiler2014visualizing}
Zeiler, M.~D., R.~Fergus.
\newblock Visualizing and understanding convolutional networks.
\newblock In \emph{Proceedings of the European Conference on Computer Vision
  (ECCV)}, pages 818--833. 2014.

\bibitem{kurakin2016adversarial}
Kurakin, A., I.~Goodfellow, S.~Bengio.
\newblock Adversarial machine learning at scale.
\newblock In \emph{International Conference on Learning Representations
  (ICLR)}. 2017.

\bibitem{zagoruyko2016wide}
Zagoruyko, S., N.~Komodakis.
\newblock Wide residual networks.
\newblock \emph{arXiv preprint arXiv:1605.07146}, 2016.

\bibitem{ribeiro2016should}
Ribeiro, M.~T., S.~Singh, C.~Guestrin.
\newblock " why should i trust you?" explaining the predictions of any
  classifier.
\newblock In \emph{Proceedings of the 22nd ACM SIGKDD international conference
  on knowledge discovery and data mining}, pages 1135--1144. 2016.

\bibitem{selvaraju2017grad}
Selvaraju, R.~R., M.~Cogswell, A.~Das, et~al.
\newblock Grad-cam: Visual explanations from deep networks via gradient-based
  localization.
\newblock In \emph{Proceedings of International Conference on Computer Vision
  (ICCV)}, pages 618--626. 2017.

\bibitem{singla2019explanation}
Singla, S., B.~Pollack, J.~Chen, et~al.
\newblock Explanation by progressive exaggeration.
\newblock In \emph{International Conference on Learning Representations
  (ICLR)}. 2020.

\bibitem{samangouei2018explaingan}
Samangouei, P., A.~Saeedi, L.~Nakagawa, et~al.
\newblock Explaingan: Model explanation via decision boundary crossing
  transformations.
\newblock In \emph{Proceedings of the European Conference on Computer Vision
  (ECCV)}, pages 666--681. 2018.

\bibitem{dosovitskiy2016generating}
Dosovitskiy, A., T.~Brox.
\newblock Generating images with perceptual similarity metrics based on deep
  networks.
\newblock In \emph{Advances in Neural Information Processing Systems
  (NeurIPS)}, pages 658--666. 2016.

\bibitem{goodfellow2014explaining}
Goodfellow, I.~J., J.~Shlens, C.~Szegedy.
\newblock Explaining and harnessing adversarial examples.
\newblock In \emph{International Conference on Learning Representations
  (ICLR)}. 2014.

\bibitem{zhou2016learning}
Zhou, B., A.~Khosla, A.~Lapedriza, et~al.
\newblock Learning deep features for discriminative localization.
\newblock In \emph{Proceedings of the IEEE Conference on Computer Vision and
  Pattern Recognition (CVPR)}, pages 2921--2929. 2016.

\bibitem{zhang2019interpreting}
Zhang, T., Z.~Zhu.
\newblock Interpreting adversarially trained convolutional neural networks.
\newblock In \emph{International Conference on Machine Learning (ICML)}. 2019.

\bibitem{etmann2019connection}
Etmann, C., S.~Lunz, P.~Maass, et~al.
\newblock On the connection between adversarial robustness and saliency map
  interpretability.
\newblock In \emph{International Conference on Machine Learning (ICML)}. 2019.

\bibitem{captum2019github}
Kokhlikyan, N., V.~Miglani, M.~Martin, et~al.
\newblock Pytorch captum.
\newblock \url{https://github.com/pytorch/captum}, 2019.

\end{thebibliography}
}

\appendix

\section{Overview}
This supplementary material provides detailed configuration information of our derivation and experiments.
Section~\ref{sec:derivation} includes a detailed derivation process of the variational lower bound for the mutual information.
Section~\ref{sec:configuration}  describes the detailed model architecture and experimental setup. Section~\ref{sec:experiment} presents additional examples of explanations.
\section{Derivation of Variational Lower Bound of Interpretability Term}\label{sec:derivation}
The derivation for Equation~3 of the paper is shown below in further detail.

To show (Eq.~3): 
\begin{equation*}
  \begin{split}
    I(G(l), f(x))
    &=I(\hat{x}, y)
    =H(y) - H(y|\hat{x})\\
    &\geq \int_{\hat{x}} p(\hat{x}) \int_y p(y|\hat{x}) \log {p(\hat{y}|\hat{x})} dy d\hat{x} + H(y)\\
    &= \mathbf{E}_{\hat{x}\sim G(l),y \sim f(x)}[\log {p(\hat{y}|\hat{x})}] + H(y) \triangleq L_I(G)
  \end{split}
\end{equation*}

Derivation: 

\begin{equation*}
  \begin{split}
    I(G(l), f(x))
    &=I(\hat{x}, y)
    =H(y) - H(y|\hat{x})\\
    &=\int_{\hat{x}}\int_y p(y, \hat{x})\log p(y|\hat{x})dyd\hat{x} + H(y)\\
    &=\int_{\hat{x}} p(\hat{x}) \int_y p(y|\hat{x}) \log p(y|\hat{x}) dyd\hat{x} + H(y)\\
    &=\int_{\hat{x}} p(\hat{x}) \int_y (p(y|\hat{x}) \log \frac{p(y|\hat{x})}{p(\hat{y}|\hat{x})} + p(y|\hat{x}) \log {p(\hat{y}|\hat{x})}) dyd\hat{x} + H(y)\\
    &=\int_{\hat{x}} p(\hat{x}) \int_y (D_{KL}(p(y|\hat{x})||p(\hat{y}|\hat{x})) + p(y|\hat{x}) \log {p(\hat{y}|\hat{x})}) dyd\hat{x} + H(y)\\
    &\geq \int_{\hat{x}} p(\hat{x}) \int_y p(y|\hat{x}) \log {p(\hat{y}|\hat{x})} dyd\hat{x} + H(y)\\
    &= \int_{\hat{x}} \int_y p(y,\hat{x}) \log {p(\hat{y}|\hat{x})} dyd\hat{x} + H(y)\\
    &= \mathbf{E}_{\hat{x}\sim G(l),y \sim f(x)}[\log {p(\hat{y}|\hat{x})}] + H(y) \triangleq L_I(G),
  \end{split}
\end{equation*}

where $D_{KL}$ is the Kullback–Leibler divergence and $H(x)$ is the entropy of random variable $x$.
\section{Model Architecture and Experiment Setup}\label{sec:configuration}
In this section, we provide additional details on the implementation, model architecture, and hyper-parameters used in this work. For the baseline methods used in Section~4, we use Captum~\cite{captum2019github} to generate explanations.

\textbf{Classification models.}
For the MNIST dataset, we use the same model architecture as used in \cite{madry2017towards, zhang2019theoretically, zhang2019you}, which has four convolutional layers and three fully-connected layers. The naturally trained model is trained for $10$ epochs with a $0.01$ learning rate in the first $5$ epochs and a $0.001$ learning rate in the last $5$ epochs.
For the CelebFaces dataset, we use VGG$16$~\cite{simonyan2014very} for gender classification.
The naturally trained model is trained for $7$ epochs with a $0.001$ learning rate.
In Section~5, for the CIFAR10 dataset, we use VGG$16$~\cite{simonyan2014very} and Wide-Resnet-34-10~\cite{zagoruyko2016wide} as the model architecture.

\textbf{Adversarial attack and adversarial training.}
For generating adversarial examples, we perform PGD-$40$ ($l_\infty, \epsilon = 0.3$)~\cite{kurakin2016adversarial} attack on the MNIST dataset,  PGD-$20$ ($l_\infty, \epsilon = 2/255$) attack on the CelebFaces dataset, and PGD-$20$ ($l_\infty, \epsilon = 8/255$) attack on 
the CIFAR10 dataset.
For adversarial training used in Section~5, we use ATTA-$1$~\cite{zheng2019efficient} with TRADES loss ($\beta = 6$)\cite{zhang2019theoretically} (and with recommended hyperparameters from ~\cite{zhang2019theoretically}) to train the model.

\textbf{Generative models.}
We use progressive GAN~\cite{karras2017progressive} to train the generative models for the MNIST~\cite{lecun1998gradient} and the CelebFaces~\cite{liu2015faceattributes} datasets. For latent variables in different layers, we use the same architecture for explanation generators. Different compression networks were used to connect latent variables and explanation generators, as described in Tables~\ref{tab:cprn-mnist}  and~\ref{tab:cprn-celeb} for MNIST and CelebFaces, respectively. We set $\lambda = 1$ for all InterpretGAN training.

For the explanation generators of the MNIST dataset, we resize the image to $32 \time 32$ for the convenience of progressive GAN training.
Explanation generators have $4$ scales, and each scale has $2$ convolutional layers and are connected with an up-sample layer. The outputs of each scale are $4 \times 4$, $8 \times 8$,
$16 \times 16$, and $32 \times 32$. The discriminators have an inversed architecture, but up-sample layers are replaced by down-sample layers.
For the first three scales, each scale is trained for $6$ epochs, and the last (fourth) scale is trained for another $12$ epochs. All epochs use a $0.001$ learning rate.
Our implementation of progressive GAN for MNIST is based on \url{https://github.com/jeromerony/Progressive_Growing_of_GANs-PyTorch}.

For the explanation generators of the CelebFaces dataset, as in progressive GAN~\cite{karras2017progressive}, we resize the image to $128 \times 128$.
Explanation generators have $6$ scales, and each scale has $2$ convolutional layers and is connected with an up-sample layer. The outputs of each scale are $4 \times 4$, $8 \times 8$,
$16 \times 16$, $32 \times 32$, $64 \times 64$, and $128 \times 128$. The discriminators have an inversed architecture but with up-sample layers replaced by down-sample layers.
Our implementation of progressive GAN is based on \url{https://github.com/facebookresearch/pytorch_GAN_zoo}.
The first scale is trained for $48,000$ iterations. For the other $5$ scales, each scale is trained for $96,000$ iterations.

\begin{table}[!htbp]
  \centering
  \begin{tabular}{cc}
    \toprule
  {\bf Layers} & {\bf Compression Network Architecture} \\ \midrule
  Conv1  &
  \tabincell{c}{$3 \times 3$ Conv, $64$ $\to$ AvgPooling $\to$
  $3 \times 3$ Conv, $128$ $\to$ \\
  AvgPooling $\to$ $3 \times 3$ Conv, $256$ $\to$ AvgPooling $\to$ FC}
  \\ \midrule
  Conv2  &
  \tabincell{c}{$3 \times 3$ Conv, $64$ $\to$ AvgPooling $\to$\\
   $3 \times 3$ Conv, $128$ $\to$ AvgPooling $\to$ FC}
  \\ \midrule
  Conv3  &
  \tabincell{c}{$3 \times 3$ Conv, $128$ $\to$ AvgPooling $\to$\\
   $3 \times 3$ Conv, $256$ $\to$ AvgPooling $\to$ FC}
  \\ \midrule
  Conv4  & FC \\ \midrule
  FC1    & FC \\ \midrule
  FC2    & FC \\ \midrule
  FC3    & FC \\ \bottomrule
  \end{tabular}
  \caption{Compression network archtectures for different layers of the MNIST model.}
  \label{tab:cprn-mnist}
\end{table}

\begin{table}[!htbp]
  \centering
  \begin{tabular}{cc}
    \toprule
  {\bf Layers} & {\bf Compression Network Architecture} \\ \midrule
  Conv Block1  &
  \tabincell{c}{
  $3 \times 3$ Conv, $128$ $\to$ AvgPooling $\to$
  $3 \times 3$ Conv, $256$ $\to$ \\
  AvgPooling $\to$ $3 \times 3$ Conv, $512$ $\to$ AvgPooling $\to$ \\
  $3 \times 3$ Conv, $512$ $\to$ AvgPooling
  }
  \\ \midrule
  Conv Block2  &
  \tabincell{c}{
  $3 \times 3$ Conv, $256$ $\to$ AvgPooling $\to$ $3 \times 3$ Conv, $512$ $\to$ \\
  AvgPooling $\to$ $3 \times 3$ Conv, $512$ $\to$ AvgPooling
  }
  \\ \midrule
  Conv Block3  &
  \tabincell{c}{
  $3 \times 3$ Conv, $512$ $\to$ AvgPooling $\to$\\
  $3 \times 3$ Conv, $512$ $\to$ AvgPooling
  }
  \\ \midrule
  Conv Block4  &
  \tabincell{c}{
    $3 \times 3$ Conv, $512$ $\to$ AvgPooling
  }
  \\ \midrule
  Conv Block5  &
  \tabincell{c}{
    $3 \times 3$ Conv, $512$
  }
  \\ \midrule
  FC1    & FC \\ \midrule
  FC2    & FC \\ \midrule
  FC3    & FC \\ \bottomrule
  \end{tabular}
  \caption{Compression network archtectures for different layers of the CelebFaces model.}
  \label{tab:cprn-celeb}
\end{table}

\section{Additional Experiment Results}\label{sec:experiment}
\begin{figure}[!htbp]
  \centering
  \includegraphics[width=1.0\textwidth]{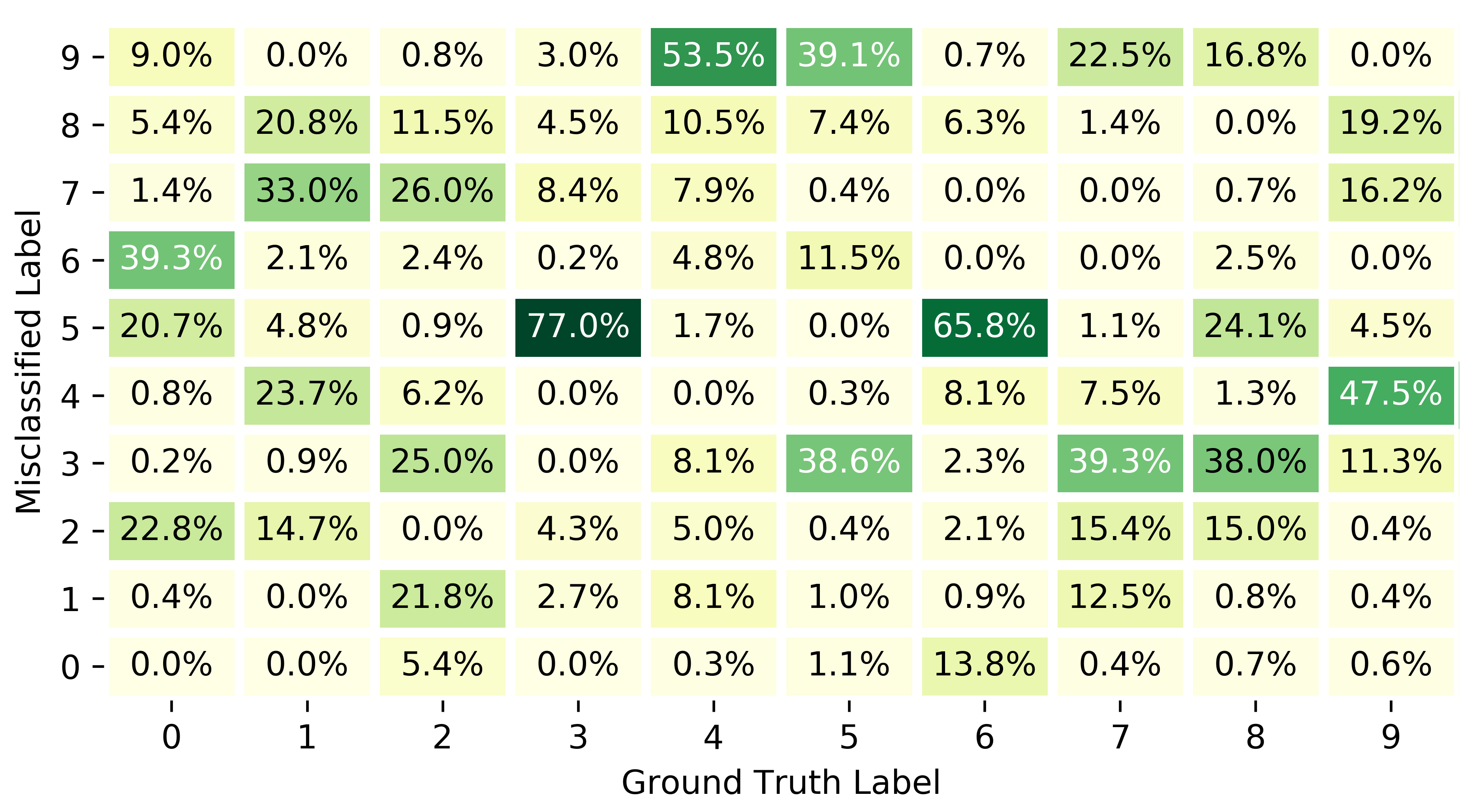}
  \caption{
    The distribution of misclassified labels: the different ground truth labels tend to be misclassified as different labels.
  }
  \label{fig:heat-map}
\end{figure}

\begin{figure}[!h]
  \centering
  \includegraphics[width=1\textwidth]{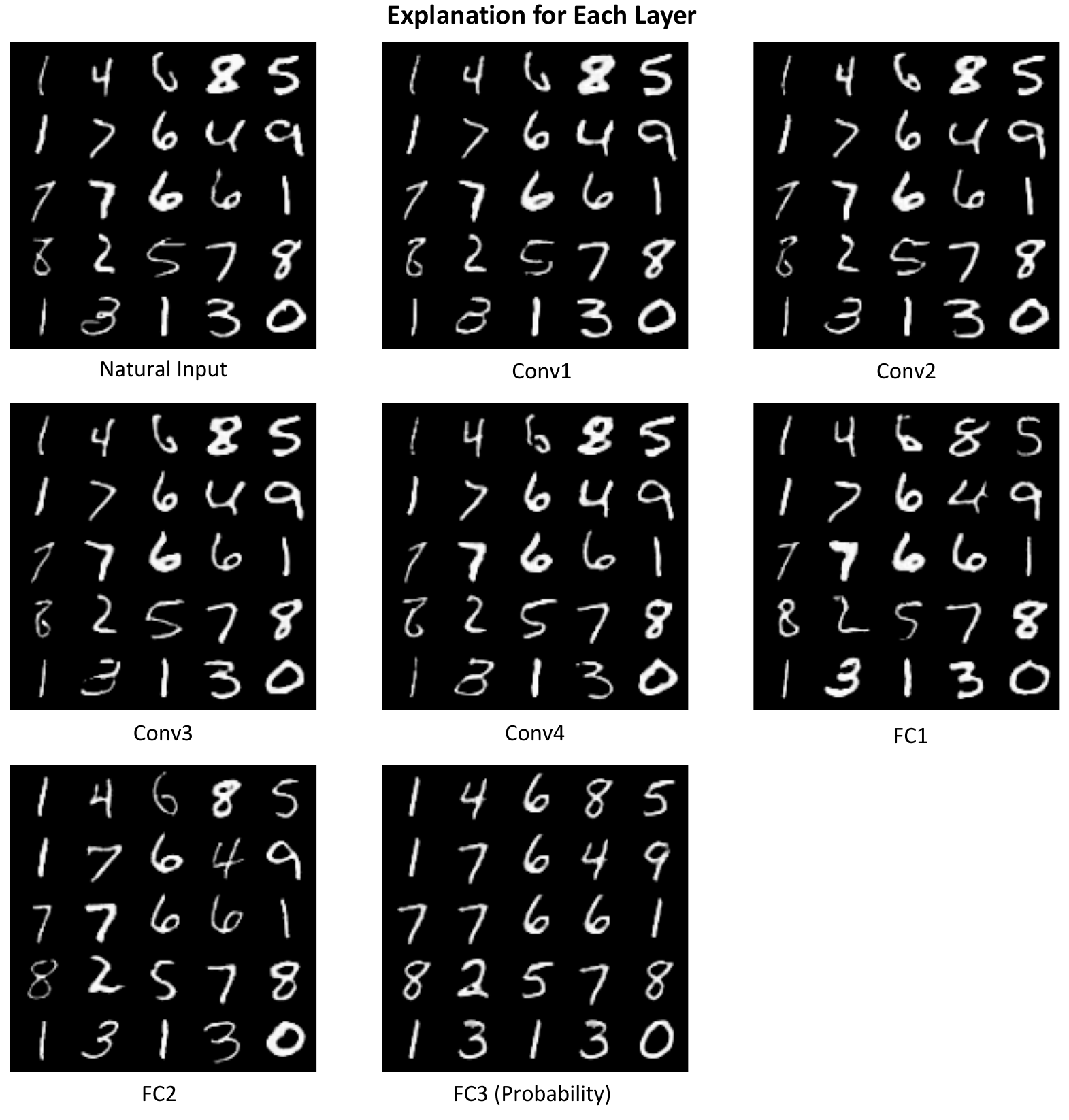}
  \caption{
    Explanations for randomly selected natural images from the test dataset.
  }
  \label{fig:nat-interpret-mnist}
\end{figure}

\begin{figure}[!h]
  \centering
  \includegraphics[width=1\textwidth]{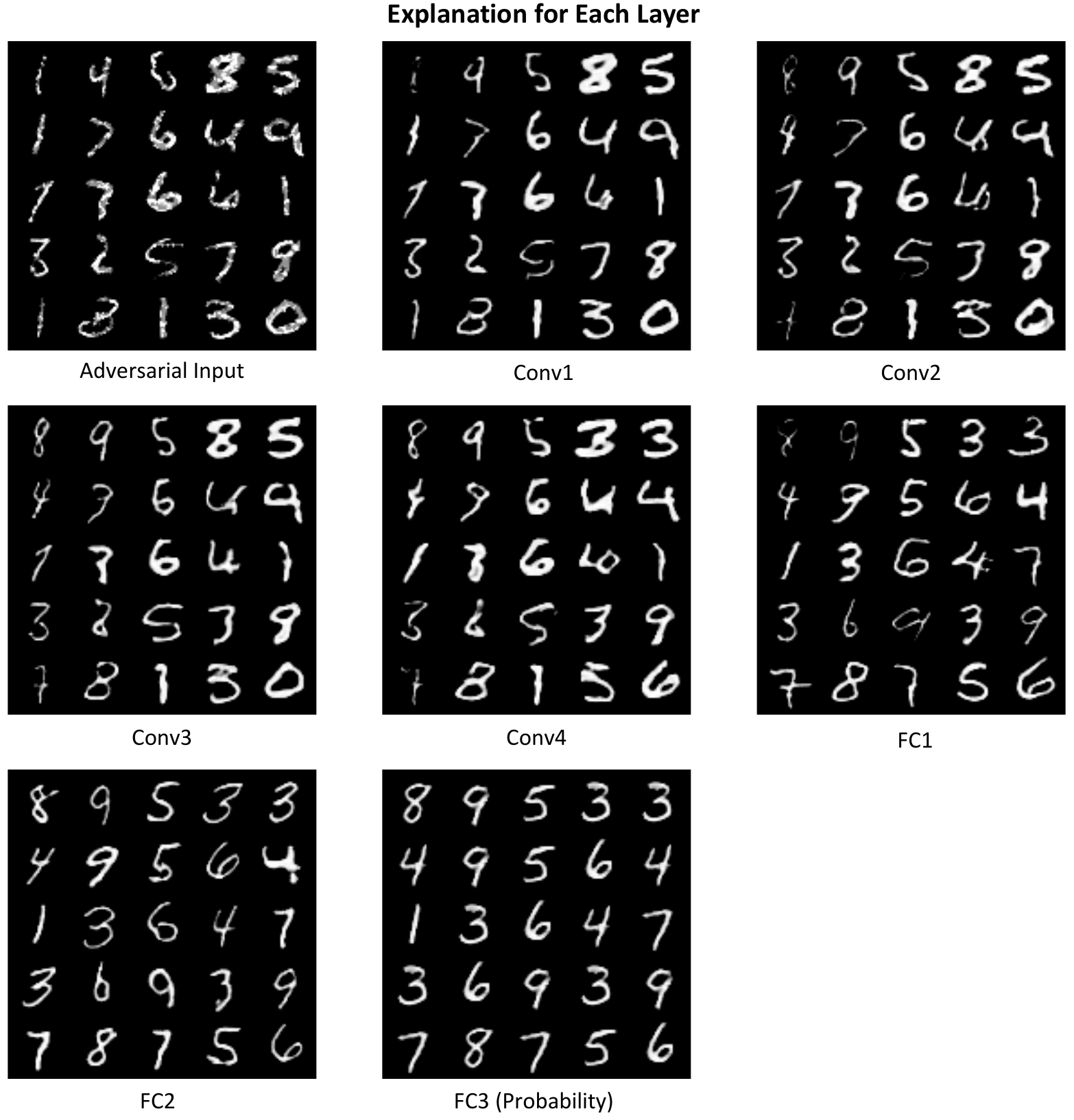}
  \caption{
    Explanations for adversarial images generated by PGD-$40$ for images in Figure~\ref{fig:nat-interpret-mnist}.
  }
  \label{fig:s-adv-interpret-mnist}
\end{figure}

In this section, we present additional examples and results for Section~4 in the paper.
Figure~\ref{fig:heat-map} shows the distribution of misclassified labels of the MNIST dataset under an untargeted attack.
As discussed in Section~4.1 in the paper, we find that, under an untargeted attack, images with different ground truth labels tend to be misclassified as different labels.
A possible reason is that, given an image with a ground truth label, it is easier for perturbations to change the image to a similar image with a misclassified label by changing the position features of some pixels, while reusing some of the original pixels.
We also present more explanation examples of classification on natural images and adversarial images in Figures~\ref{fig:nat-interpret-celeb}, \ref{fig:nat-interpret-mnist},  \ref{fig:s-adv-interpret-mnist}, and \ref{fig:s-adv-interpret-celeb}.

\begin{figure}[!h]
  \centering
  \includegraphics[width=1\textwidth]{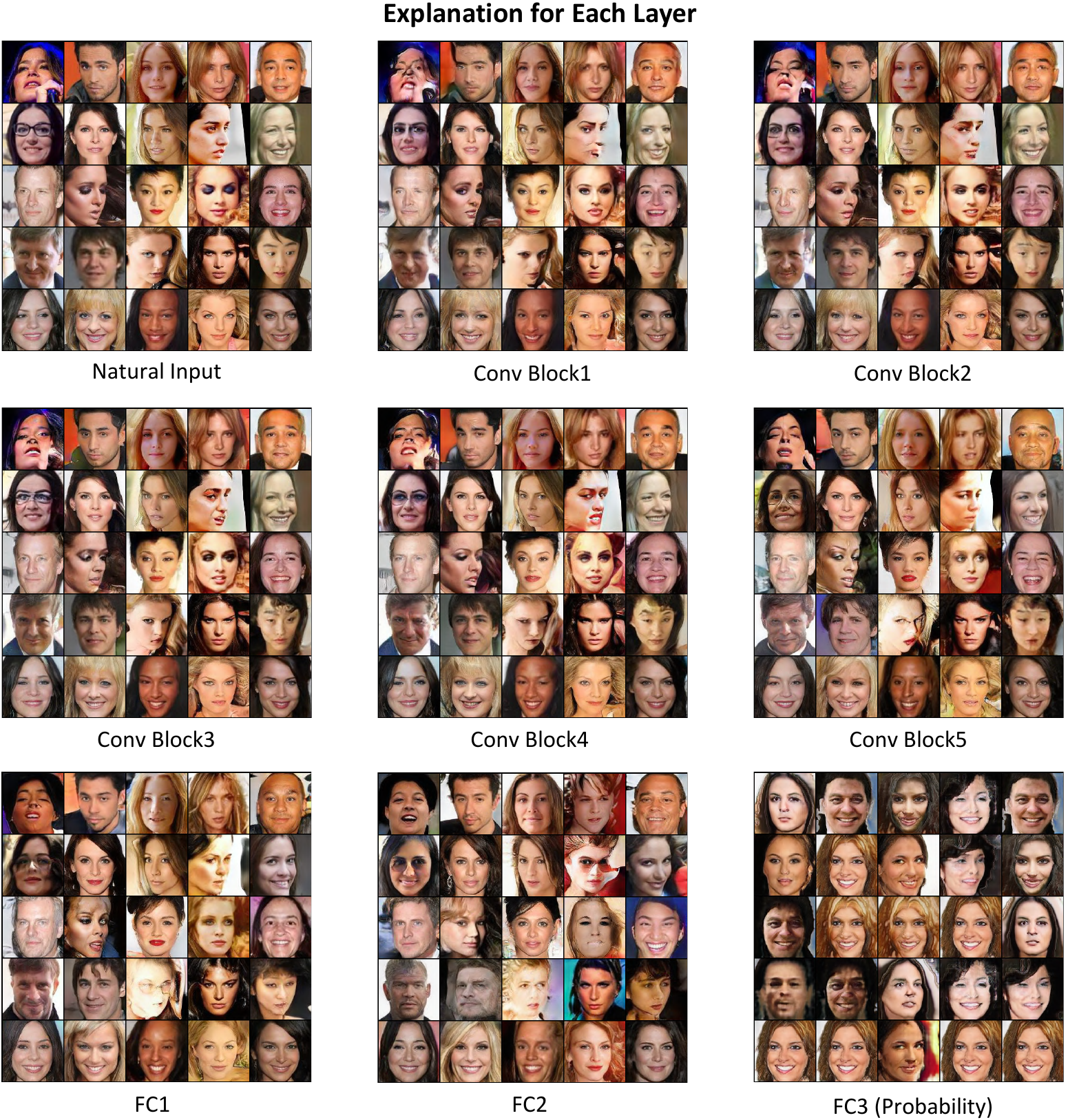}
  \caption{
    Explanations for randomly selected natural images from the test dataset.
  }
  \label{fig:s-nat-interpret-celeb}
\end{figure}

\begin{figure}[!h]
  \centering
  \includegraphics[width=1\textwidth]{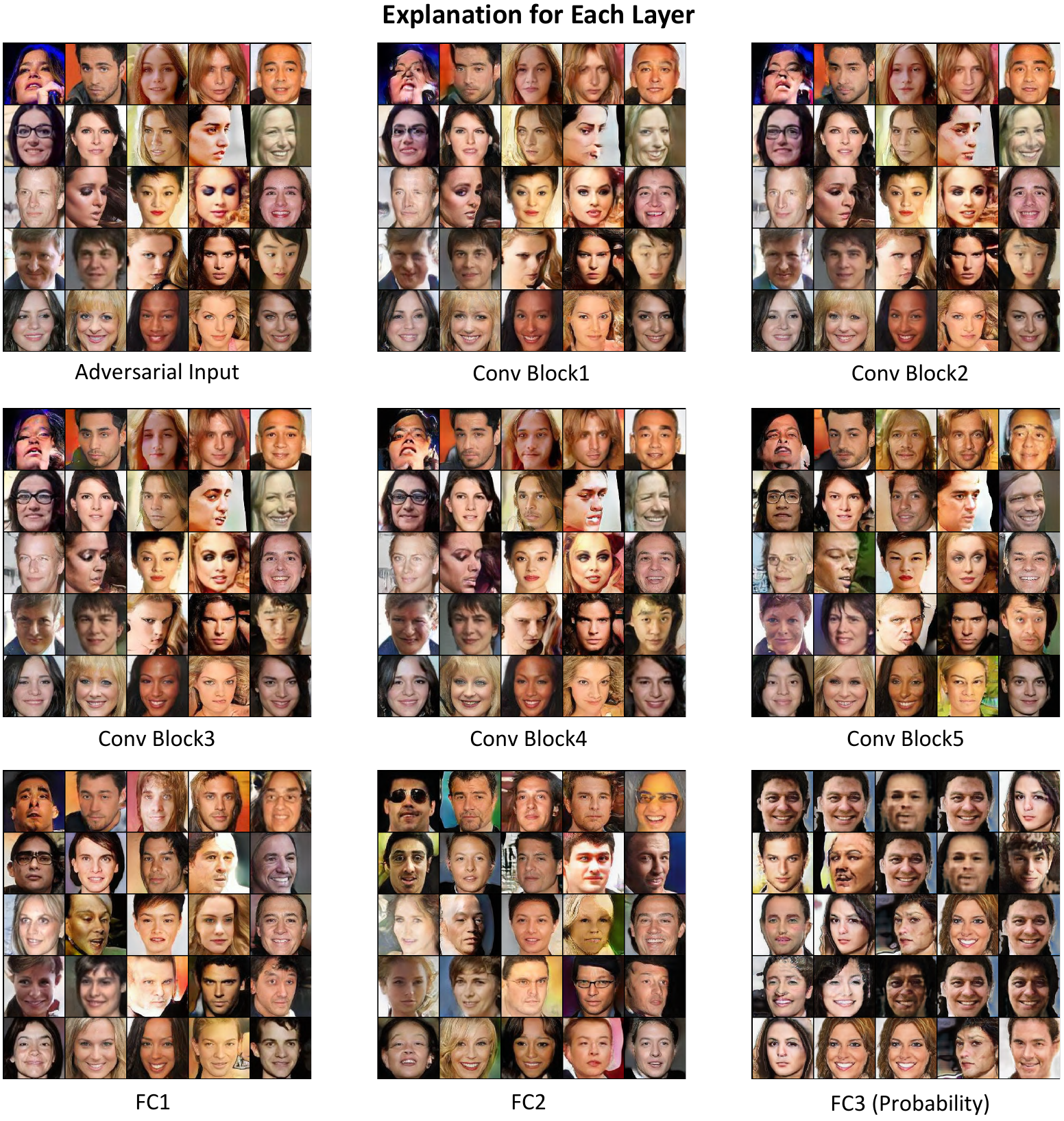}
  \caption{
    Explanations for adversarial images generated by PGD-$40$ for images in Figure~\ref{fig:s-nat-interpret-celeb}.
  }
  \label{fig:s-adv-interpret-celeb}
\end{figure}

\end{document}